\documentclass[letterpaper,10pt]{article}

\usepackage[preprint]{probml}

\ShortHeadings{CIRCUS: Circuit Consensus under Uncertainty via Stability Ensembles}

\usepackage{mathtools}
\usepackage{amssymb}
\usepackage{booktabs}
\usepackage{array}
\usepackage{graphicx}
\usepackage{multirow}
\usepackage{microtype}

\begin{document}

\title{CIRCUS: Circuit Consensus under Uncertainty\\via Stability Ensembles}

\author[1]{Swapnil Parekh}
\affil[1]{Intuit}

\maketitle

\begin{abstract}
Every mechanistic circuit carries an invisible asterisk: it reflects not just the model's
computation, but the analyst's choice of pruning threshold.
Change that choice and the circuit changes, yet current practice treats a single
pruned subgraph as ground truth with no way to distinguish robust structure from
threshold artifacts.
We introduce \textbf{CIRCUS}, which reframes circuit discovery as a problem of
\emph{uncertainty over explanations}.
CIRCUS prunes one attribution graph under $B$ configurations, assigns each edge an
empirical inclusion frequency $s(e)\in[0,1]$ measuring how robustly it survives across
the configuration family, and extracts a \emph{consensus circuit} of edges present in
every view.
This yields a principled core/contingent/noise decomposition (analogous to
posterior model-inclusion indicators in Bayesian variable selection) that separates
robust structure from threshold-sensitive artifacts, with negligible overhead.
On Gemma-2-2B and Llama-3.2-1B, consensus circuits are ${\sim}40\times$ smaller
than the union of all configurations while retaining comparable influence-flow
explanatory power, consistently outperform influence-ranked and random baselines,
and are confirmed causally relevant by activation patching.
\end{abstract}

\section{Introduction}\label{sec:intro}

Identifying \emph{circuits} (sparse causal subgraphs) is a central goal of mechanistic
interpretability \citep{olah2022mechinterp,rai2024practical}.
Modern pipelines construct \emph{attribution graphs}, directed graphs whose nodes are
learned features (produced by cross-layer transcoders, CLTs) and whose weighted edges
represent direct causal effects between features, then prune by cumulative influence to
obtain a compact circuit \citep{ameisen2025methods,lindsey2025biology}.
A critical but often unstated issue is that the result depends on arbitrary analyst
choices: two different pruning thresholds yield two different circuits, with no principled
way to tell which edges reflect genuine model structure versus threshold sensitivity.
This is, at its core, an \emph{uncertainty quantification} problem: the analyst faces a
family of plausible explanations and must decide which structural claims are robust.

\begin{figure}[t]
  \centering
  \vspace{-24pt}
  \includegraphics[width=0.75\linewidth]{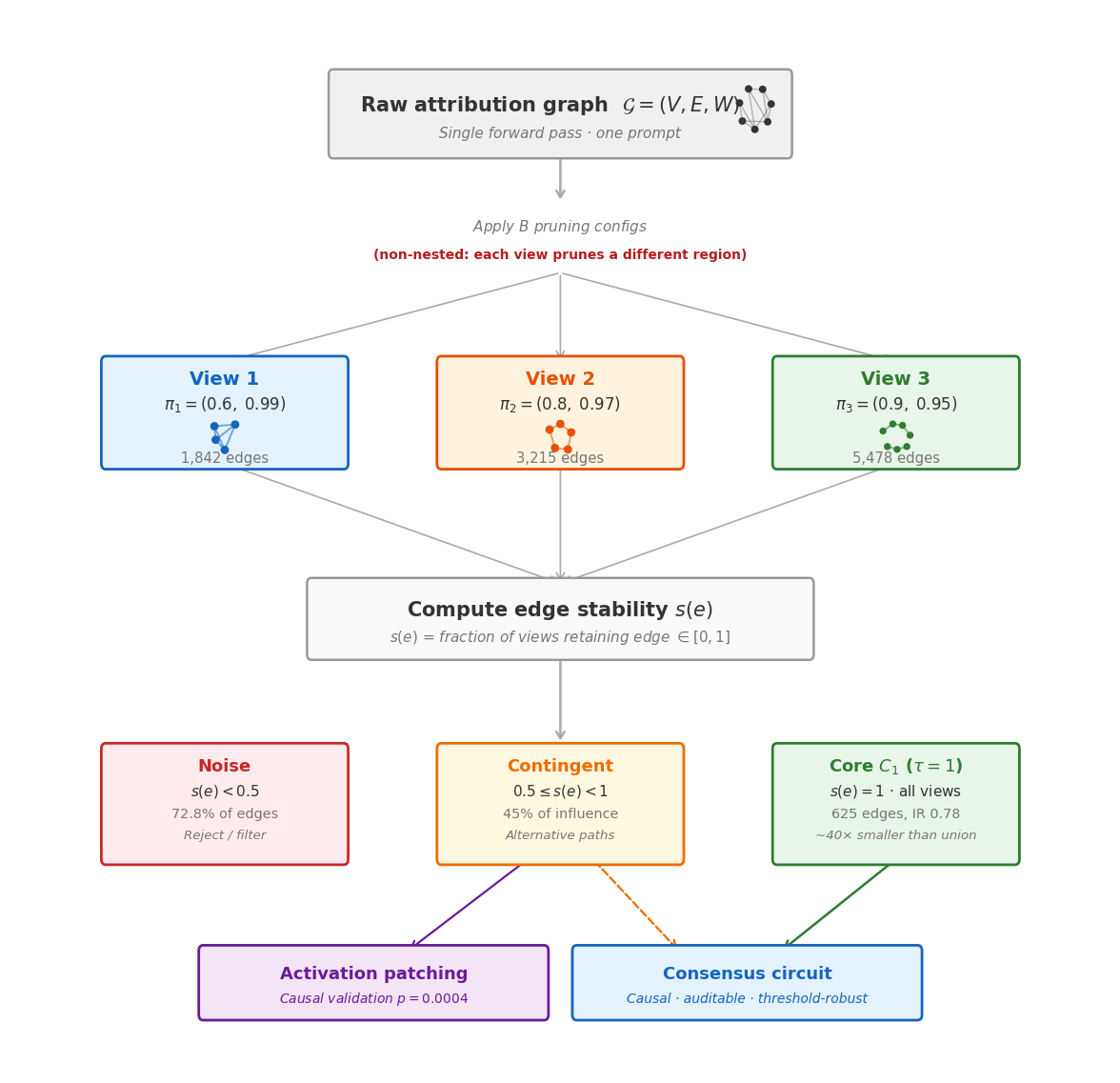}
  \vspace{-10pt}
  \caption{\textbf{CIRCUS pipeline.} A single attribution graph is pruned under $B$
    configurations to yield multiple views; edges receive stability scores $s(e)$.
    Strict consensus $C_{\tau=1}$ keeps only edges present in all views (solid lines);
    dashed edges are contingent alternatives.
    The core/contingent/noise taxonomy is described in \cref{sec:method}.}
  \label{fig:pipeline}
\end{figure}

We address this by adapting \emph{stability selection} \citep{meinshausen2010stability}
to circuit discovery.
Rather than subsampling data, CIRCUS varies pruning \emph{configurations} to generate
$B$ views of the same attribution graph, assigns each edge an empirical inclusion
frequency $s(e) \in [0,1]$ reflecting how often it survives across that configuration
family, and extracts a \emph{consensus circuit} of edges present in every view
(\cref{fig:pipeline}).
The resulting per-edge scores play a role analogous to posterior inclusion probabilities
in Bayesian variable selection: they summarize which
structural claims are supported across plausible analysis choices, enabling the analyst
to \emph{abstain} from reporting unstable edges.
A key design principle is that configurations must be \emph{non-nested}: if one config's
pruned set is always a superset of another's, the consensus degenerates to the
intersection of all views, which equals the tightest single config (see
\cref{sec:config-bagging} for a formal statement).
Crossing node and edge thresholds in opposite directions ensures consensus is a
genuinely new set that cannot be recovered from any single view.
The pipeline adds 5.5\% overhead beyond a single attribution run (\cref{sec:results}).

\noindent\textbf{Contributions.}
\textbf{(Method)}~A config-bagging pipeline that turns analyst-choice variability into an
uncertainty estimate: one attribution run, $B$ non-nested configurations,
frequency-based inclusion scores $s(e)\in[0,1]$, and a core/contingent/noise taxonomy.
\textbf{(Empirics)}~On Gemma-2-2B (50 prompts) and Llama-3.2-1B (20 prompts)
\citep{gemma2024,meta2024llama}, consensus is ${\sim}40\times$ smaller than the union
at comparable IR and outperforms influence-ranked and random baselines on most prompts
(\cref{tab:mve}).
\textbf{(Faithfulness)}~$400\times$ lower mean KL divergence vs.\ union-pruned.
\textbf{(Validation)}~Activation patching \citep{zhang2024activation} confirms causal
relevance over matched controls ($p{=}0.0004$, $57\%$ of oracle recovery).

\section{Related Work}\label{sec:related}

\textbf{Stability selection and ensembling.}
Our method adapts \emph{stability selection} \citep{meinshausen2010stability},
replacing data subsampling with \emph{config-bagging} over pruning thresholds.
The original framework provides FDR control under exchangeability; our deterministic
configuration family does not satisfy that assumption, so $s(e)$ is an empirical
robustness score analogous to posterior inclusion probabilities in Bayesian variable
selection, though without their formal guarantees.
We chose counting over a fully Bayesian treatment because the discrete, low-dimensional
threshold grid makes posterior inference unnecessary.
\citet{gadgil2025ensembling} ensemble SAE features; \citet{gyawali2022ensembling}
ensemble feature importances; unlike these, we ensemble \emph{structure} (which edges
appear), not saliency magnitudes \citep{paillard2026aggregate}.

\textbf{Uncertainty in circuits.}
\citet{krasnovsky2025eics} study circuit uncertainty via coherence scores;
\citet{bley2024explaining} explain predictive uncertainty via second-order attribution;
\citet{agarwal2022rethinking} quantify attribution stability under input perturbation.
We focus on uncertainty over \emph{explanation structure} when analyst choices vary,
yielding a rejection criterion for unreliable edges \citep{zhu2025robust}.

\section{Method}\label{sec:method}

\subsection{Notation and Problem Formulation}

Let $\mathcal{G} = (V, E, W)$ denote the \emph{full attribution graph} for a fixed prompt
and target token: $V$ is the set of nodes (features, errors, tokens, logits), $E$ the
directed edge set, and $W \in \mathbb{R}^{|E|}$ the edge weights (direct effects),
produced by a CLT replacement model \citep{ameisen2025methods,circuit-tracer}.
A \emph{pruning configuration} $\pi$ (node and edge cumulative-influence thresholds)
yields a pruned edge set $E^{(\pi)} \subseteq E$.
We treat $B$ configurations as \emph{views}: the $b$-th view has edge set $E^{(b)}$.

\textbf{Edge stability.}
For each edge $e$, the \emph{stability score} is its empirical selection probability
over the design set of configurations:
\begin{equation}\label{eq:stability}
  s(e) \;=\; \frac{1}{B} \sum_{b=1}^{B} \mathbb{1}[e \in E^{(b)}].
\end{equation}
$s(e) \in [0,1]$; $s(e)=1$ iff $e$ appears in every view.
$s(e)$ is a \emph{conditional} frequency: it measures robustness across the chosen
configuration family $\{\pi_b\}$, not across all possible pruning strategies.
The score is inspired by stability selection \citep{meinshausen2010stability} but carries
no formal FDR guarantee, as our design set is deterministic and not exchangeable.
Conceptually, $s(e)$ serves as a frequentist analogue of a posterior inclusion
probability: it reports the fraction of plausible analysis configurations under which
edge $e$ is retained, without claiming a probabilistic generative model over
configurations.

\textbf{Consensus.}
\begin{equation}\label{eq:consensus}
  C_\tau \;=\; \bigl\{\, e : s(e) \geq \tau \,\bigr\}.
\end{equation}
We use $\tau{=}1$ for the strict consensus (all-views circuit) and $\tau{<}1$ for a larger
exploratory set.
The threshold $\tau$ controls a \emph{stability--coverage tradeoff}: higher $\tau$ yields
a smaller, more robust circuit at the cost of lower influence retained
(\cref{fig:elbow}).
The \emph{union} is $E^{\mathrm{union}} = \bigcup_b E^{(b)}$.

\textbf{Influence retained (IR).}
\begin{equation}\label{eq:ir}
  \mathrm{IR}(S) \;=\; \frac{\Phi(S)}{\Phi(E)},
\end{equation}
where $\Phi(S) = \sum_{e \in S} |w_e|$ sums absolute edge weights (direct effects) in
$S$, and $\Phi(E)$ is the same sum over all edges in the full attribution graph
\citep{ameisen2025methods}.
$\mathrm{IR}(S) \in [0,1]$ measures what fraction of total attributed influence passes
through $S$; it is a structural proxy for explanatory coverage, not a causal
measure. We complement it with activation patching in \cref{sec:results}.

\subsection{Config-Bagging and Circuit Taxonomy}\label{sec:config-bagging}

We address analyst-choice variability by treating plausible pruning thresholds as a
finite design set (\textbf{config-bagging}): one attribution run, $B$ configurations,
$B$ views, no retraining.
$s(e)$ quantifies robustness to threshold choice within the chosen family;
multi-prompt IR variance separately captures instance-level variability.

\noindent\textbf{Non-nesting principle.}
If configurations are nested ($E^{(1)} \subseteq E^{(2)} \subseteq \cdots \subseteq
E^{(B)}$), then $C_{\tau=1} = E^{(1)}$: consensus simply recovers the tightest single
view.
We therefore require \emph{non-nested} configurations, achieved by crossing node and
edge thresholds in opposite directions so that each view prunes a different part of
the graph (\cref{sec:nesting}).

\noindent\textbf{Circuit taxonomy.}
We partition union edges into three tiers:
\emph{Core}: $s(e){=}1$ (present in every view), forming $C_1$.
\emph{Contingent}: $0.5 \leq s(e) < 1$, edges in at least half of views that are
alternative pathways worth reporting alongside the core.
\emph{Noise}: $s(e) < 0.5$, edges in fewer than half of views, flagged for rejection.
This \emph{abstention} capability connects CIRCUS to selective-prediction ideas:
the analyst can restrict explanations to edges stable across the configuration family.
If $C_1$ has low IR, a residual \textbf{boosting} round constructs $C_2$ from the
residual graph, yielding a \emph{boosted} circuit $C_1 \cup C_2$; see
\cref{apd:boosting}.

\section{Experiments and Results}\label{sec:results}

\textbf{Setup.}
We use Gemma-2-2B and Llama-3.2-1B \citep{gemma2024,meta2024llama} with publicly
available CLTs \citep{ameisen2025methods,circuit-tracer}.
Prompts are short factoid completions (capitals, arithmetic, trivia); generalization to
more complex behaviors (reasoning, in-context learning) remains open.
We use $B{=}25$ (a $5{\times}5$ grid) for stability-distribution analysis and $B{=}9$
non-nested configs for the main win-rate evaluation (\cref{sec:nesting}).
We evaluate on 50 prompts (Gemma) and 20 prompts (Llama).
\textbf{Baselines} (all at the same edge budget $|C_1|$):
(1)~\emph{union-pruned}: top-$|C_1|$ union edges ranked by influence;
(2)~\emph{random}: $|C_1|$ edges drawn uniformly from the union (averaged over 10
seeds).
The full pipeline (attribution + $B$ prunes + consensus + IR) takes 0.19s beyond the
single attribution run (5.5\% overhead), with the consensus build itself
taking $<1$ms.

\textbf{Stability--coverage tradeoff ($B{=}25$).}
\cref{fig:elbow} shows the tradeoff: as $\tau$ increases, $|C_\tau|$ drops while IR
remains high until strict consensus, producing an elbow (bootstrap 95\% CI at
$\tau{=}1$: $|C_1| \in [180, 265]$, $\mathrm{IR}(C_1) \in [0.72, 0.74]$).
Mean per-edge influence increases monotonically with stability: edges surviving all 25
configs carry ${\sim}70\times$ higher mean influence than single-config edges
(\cref{apd:stability-inf}).

\textbf{Addressing config nesting.}\label{sec:nesting}
As predicted by the non-nesting principle (\cref{sec:config-bagging}), nested configs
$(0.6, 0.95) \subset (0.8, 0.98) \subset (0.9, 0.99)$ yield
consensus trivially equal to the loosest configuration (Match = 50/50 prompts).
Non-nested configs resolve this artifact: e.g.,
$(0.6, 0.99)$ retains few nodes but many edges, while $(0.9, 0.95)$ does the opposite.
\cref{tab:mve} shows that increasing $B$ with non-nested configs resolves the nesting
artifact while improving win rates against both baselines.

\begin{table}[t]
  \centering
  \caption{Wide-scale results: mean IR and win rates across prompts (Gemma-2-2B, 50
    prompts; Llama-3.2-1B, 20 prompts). Non-nested configs cross node/edge thresholds.
    All baselines use $|C_1|$ edges. ``Match'' = consensus equals a single config
    (Match $= B/B$ indicates the consensus is trivially the tightest view, a nesting
    artifact; lower Match values confirm that consensus is a genuinely new circuit).}
  \label{tab:mve}
  \small
  \begin{tabular}{@{}llcccc@{}}
    \toprule
    \textbf{Model} & \textbf{Configs} & \textbf{IR(C$_1$)} & \textbf{W-UP} & \textbf{W-Rand} & \textbf{Match} \\
    \midrule
    \multirow{3}{*}{Gemma}
      & nested $B{=}3$      & $.817{\pm}.045$ & 50/50 & 37/50 & 50/50 \\
      & non-nest.\ $B{=}5$  & $.822{\pm}.044$ & 30/50 & 20/50 & 3/50  \\
      & non-nest.\ $B{=}9$  & $.808{\pm}.051$ & \textbf{47/50} & \textbf{37/50} & 8/50  \\
    \midrule
    \multirow{3}{*}{Llama}
      & nested $B{=}3$      & $.886{\pm}.044$ & 20/20 & 20/20 & 20/20 \\
      & non-nest.\ $B{=}5$  & $.883{\pm}.044$ & 16/20 & 20/20 & 0/20  \\
      & non-nest.\ $B{=}9$  & $.892{\pm}.052$ & \textbf{19/20} & \textbf{20/20} & 0/20  \\
    \bottomrule
  \end{tabular}
\end{table}

\textbf{View diversity and rejection.}
With non-nested $B{=}9$, mean pairwise Jaccard is 0.33 (Gemma) and 0.35 (Llama),
confirming genuine view diversity; leave-one-config-out analysis confirms no single
configuration drives the consensus.

\begin{figure}[t]
  \centering
  \begin{minipage}[t]{0.38\linewidth}
    \centering
    \includegraphics[width=\linewidth]{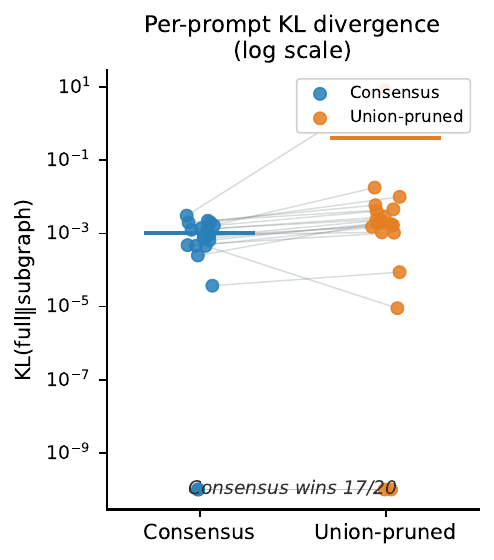}
  \end{minipage}\hfill
  \begin{minipage}[t]{0.58\linewidth}
    \centering
    \includegraphics[width=\linewidth]{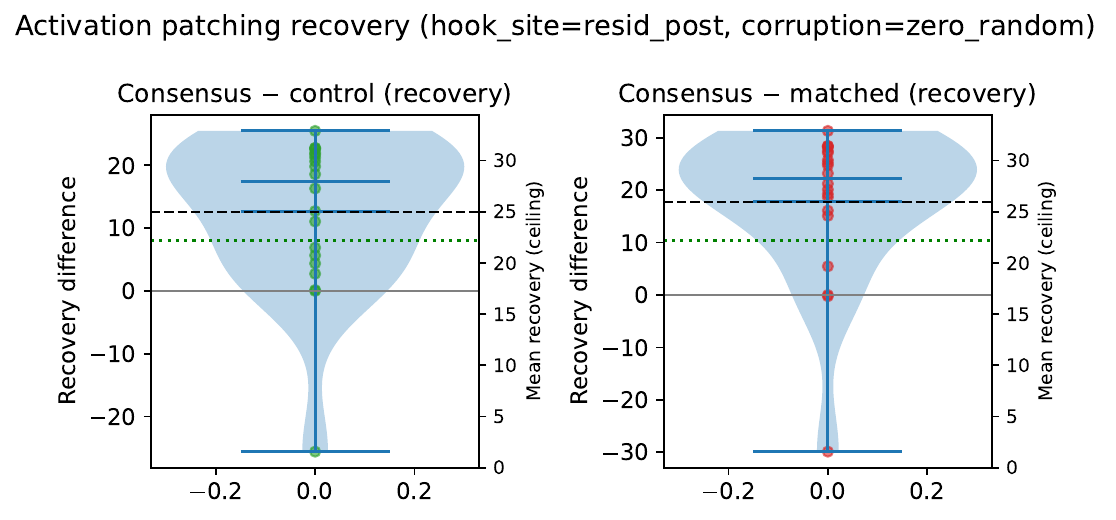}
  \end{minipage}
  \caption{\textbf{Left:} Per-prompt KL divergence (log scale); consensus (blue) clusters
    near $10^{-3}$ while union-pruned (orange) collapses on outliers.
    \textbf{Right:} Activation patching recovery; consensus at $57\%$ of oracle
    ($p{=}0.0004$) vs.\ $20\%$ for random.}
  \label{fig:results}
\end{figure}

\textbf{Causal validation.}
We validate via activation patching \citep{zhang2024activation}: patch $C_1$ node
activations from a source prompt (different answer) to a target prompt and measure
output-logit shift toward the source answer (``recovery'').
Three conditions on $n{=}20$ prompts: \emph{consensus} ($C_1$ nodes),
\emph{random} (equal-count random nodes), and \emph{matched}
(equal-count nodes matched on influence).
Mean recovery: $19.4$ (consensus) vs.\ $6.8$ (random) and $1.6$ (matched),
$p{=}0.0004$ by Wilcoxon signed-rank test.
The oracle ceiling (all union nodes) is 34.2, placing consensus at $57\%$
of oracle vs.\ $20\%$ for random (\cref{fig:results}).

\textbf{Multi-prompt baselines and cross-model replication.}
\cref{tab:mve} summarizes the wide-scale evaluation.
With non-nested $B{=}9$, consensus matches a single config in only 8/50 (Gemma) and
0/20 (Llama) prompts, confirming genuine multi-view aggregation; increasing $B$ from 3
to 9 resolves the nesting artifact while win rates remain high.
Alternative aggregation rules do not improve over strict consensus $C_1$
(\cref{apd:aggregation}).

\textbf{Faithfulness beyond IR.}
We compare per-logit influence distributions via
$D_{\mathrm{KL}}(p_E \| p_S)$, where $p_S(\ell)$ is the fraction of subgraph
influence attributed to output logit $\ell$ and $p_E(\ell)$ the same for the full graph.
Consensus achieves $0.0010{\pm}0.0008$ vs.\ $0.40{\pm}1.74$ for union-pruned across
20 prompts (17/20 wins; \cref{fig:results}).
The large union-pruned variance reflects heavy-tailed outliers from greedy influence
ranking, which concentrates edges on the highest-weight logits; the gap is robust to
choice of summary statistic.

\section{Conclusion}\label{sec:discussion}

Circuit discovery has a measurement problem: different analyst choices yield different circuits, with no principled way to separate robust structure from threshold-sensitive artifacts.
CIRCUS addresses this by computing empirical inclusion frequencies over a configuration family, turning analyst-choice variability into a diagnostic.
The resulting core/contingent/noise taxonomy makes analyst disagreement actionable,
and demonstrates that ideas from probabilistic model selection transfer naturally to
mechanistic interpretability.

\textbf{Limitations.}
$s(e)$ is a conditional frequency over the chosen configuration family, not a posterior over all pruning strategies, so the stability--coverage tradeoff will shift under different families.
Our experiments are also limited to 1--2B models on short factoid prompts, and whether the stability structure generalizes to larger models or more complex behaviors remains an open question.

\noindent\textbf{Generative AI.} Generative AI was used for programming support, editing, and polishing of the manuscript. The author reviewed and takes full responsibility for the content.

\subsection*{Acknowledgements}
The author thanks the teams that open-sourced attribution tools and public transcoder checkpoints.

\bibliography{circus_refs}

\clearpage
\appendix

\section{Boosting Results}\label{apd:boosting}

After extracting $C_1$ (core, 625 edges, IR 0.78 with nested $B{=}3$), a residual
boosting round adds $C_2$ from the top-90\% residual influence.
\cref{tab:boosting} reports tiered results.
The full circuit $C_1 \cup C_2$ achieves IR 0.96 at 36,350 edges.
The compact variant (9,546 edges, IR 0.91) and post-pruned circuit (18,690 edges, IR
0.94) give better size--coverage trade-offs.

\begin{table}[h]
  \centering
  \caption{Tiered boosting: core, residual, full, and variants ($B{=}3$).}
  \label{tab:boosting}
  \begin{tabular}{lrrr}
    \toprule
    \textbf{Circuit} & \textbf{Edges} & \textbf{Nodes} & \textbf{IR} \\
    \midrule
    $C_1$ (core)              & 625    & 61    & 0.78 \\
    $C_2$ (residual)          & 35,725 & 1,183 & ---  \\
    $C_1 \cup C_2$ (full)     & 36,350 & 1,184 & 0.96 \\
    $C_1 \cup C_2$ compact    & 9,546  & 942   & 0.91 \\
    Post-pruned (95\%)        & 18,690 & 1,045 & 0.94 \\
    \bottomrule
  \end{tabular}
\end{table}

\section{Stability vs.\ Influence}\label{apd:stability-inf}

\begin{figure}[h]
  \centering
  \includegraphics[width=0.70\linewidth]{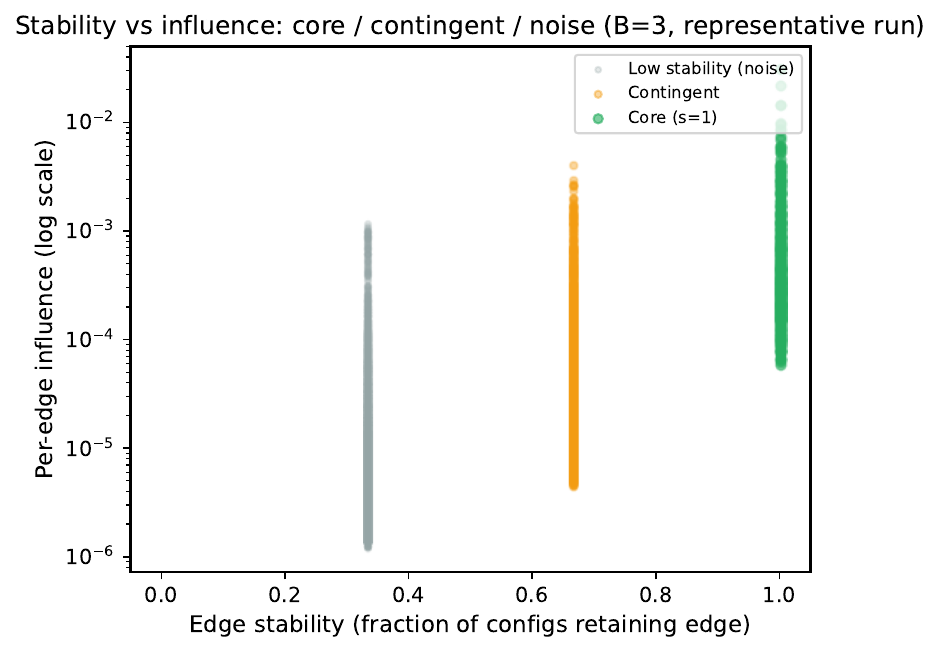}
  \caption{\textbf{Stability vs.\ influence.}
    Mean influence per edge (log scale) rises monotonically with stability score;
    edges surviving all 25 configurations carry ${\sim}70\times$ higher mean influence
    than single-configuration edges, within the chosen $5{\times}5$ grid family.}
  \label{fig:stability-inf}
\end{figure}

\section{Stability--Coverage Tradeoff}\label{apd:coverage}

\begin{figure}[h]
  \centering
  \includegraphics[width=0.80\linewidth]{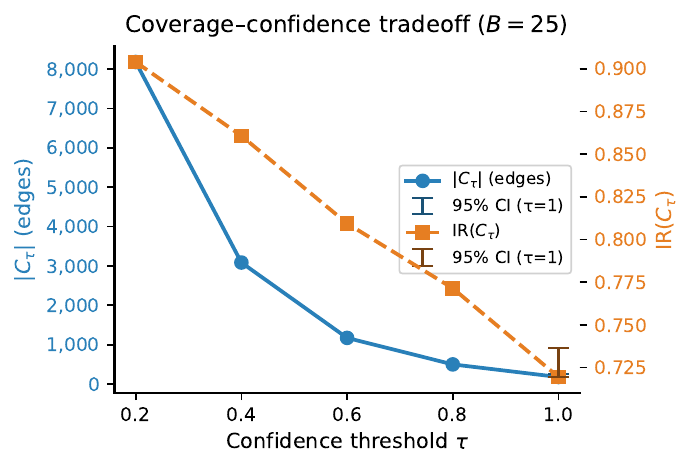}
  \caption{\textbf{Stability--coverage tradeoff} ($B{=}25$ configs).
    As the stability threshold $\tau$ increases, circuit size shrinks (left axis)
    and IR decreases gracefully (right axis).
    Error bars at $\tau{=}1$ are bootstrap 95\% CIs.
    This curve is specific to the chosen configuration family; the tradeoff shifts
    with different families.}
  \label{fig:elbow}
\end{figure}

\section{Leave-One-Out Consistency}\label{apd:calibration}

\begin{figure}[h]
  \centering
  \includegraphics[width=0.70\linewidth]{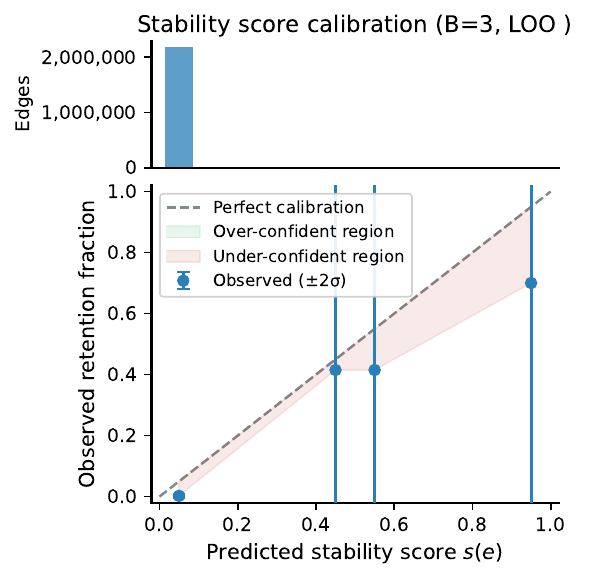}
  \caption{\textbf{Leave-one-out consistency of $s(e)$} ($B{=}3$).
    Top panel: histogram of edge counts per stability bin.
    Bottom panel: stability score $s(e)$ vs.\ observed retention fraction in the
    held-out configuration; all three points lie on the diagonal, showing that
    $s(e)$ is a consistent empirical frequency within this configuration family.}
  \label{fig:calibration}
\end{figure}

\section{Bootstrap Robustness of Strict Consensus}\label{apd:bootstrap}

\begin{figure}[h]
  \centering
  \includegraphics[width=0.80\linewidth]{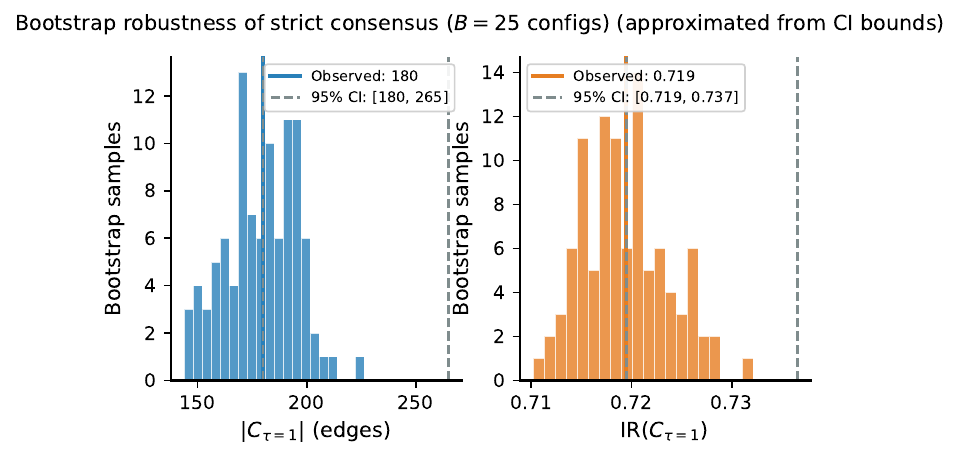}
  \caption{\textbf{Bootstrap distribution of strict-consensus size and IR} ($B{=}25$).
    Histograms show 100 bootstrap samples (config resampling with replacement) for
    $|C_{\tau=1}|$ (left) and $\mathrm{IR}(C_{\tau=1})$ (right), with 95\% CIs.
    The tight distributions confirm that the consensus circuit is robust to which
    specific $B$ configs are sampled.}
  \label{fig:bootstrap}
\end{figure}

\section{Uncertainty Analyses Summary}\label{apd:uncertainty}

\begin{table}[h]
  \centering
  \caption{Full uncertainty analyses summary ($B{=}3$, $n{=}20$ prompts).}
  \label{tab:uncertainty}
  \small
  \begin{tabular}{@{}l@{\quad}>{\raggedright\arraybackslash}p{7cm}@{}}
    \toprule
    \textbf{Analysis} & \textbf{Summary} \\
    \midrule
    Rejection           & 2.5\% stability$=1$ (625 edges); 73\% stability$<0.5$. \\
    Alternatives        & 6.2k contingent edges carrying 45\% of total influence. \\
    Approx.\ intervention & Spearman 0.29 (stability, ablation effect). \\
    Ablation ($n{=}20$) & 16/20 vs.\ matched; diff 6.0 [2.7, 9.2], $p{=}0.012$. \\
    Patching ($n{=}20$) & 17/20 vs.\ matched; diff 17.8 [10.8, 23.1], $p{=}0.0026$. \\
    Prompt-level        & Spearman(IR, prob) $= 0.69$ [0.28, 0.89]. \\
    CI width (mean IR)  & Consensus 0.074; single-config 0.103. \\
    Sensitivity         & $B{=}2,3$: 625 edges, IR 0.78; alt.\ triple: 887, IR 0.79. \\
    \bottomrule
  \end{tabular}
\end{table}

\section{Algorithm: Stability and Consensus}\label{apd:algorithm}

\noindent\textbf{Input:} Full graph $\mathcal{G}{=}(V,E,W)$; $B$ configurations
$\{\pi_b\}$; threshold $\tau \in [0,1]$.
\noindent\textbf{Output:} Per-edge stability $s(e)$; consensus $C_\tau$; union
$E^{\mathrm{union}}$.

\begin{enumerate}
  \item For each $b \in [B]$, apply $\pi_b$ to $\mathcal{G}$ to obtain $E^{(b)}$.
  \item For each $e \in E^{\mathrm{union}} = \bigcup_b E^{(b)}$, compute
    $s(e) = \frac{1}{B}\sum_{b=1}^B \mathbb{1}[e \in E^{(b)}]$.
  \item Set $C_\tau = \{e : s(e) \geq \tau\}$.
  \item Optionally compute $\mathrm{IR}(C_\tau)$ via \cref{eq:ir}.
\end{enumerate}

\noindent\textbf{Boosting:} Zero $C_1$ edges in $\mathcal{G}$; compute residual
per-edge influence; let $C_2$ capture the top-$\alpha$ fraction; return $C_1 \cup C_2$.

\section{Aggregation Rule Comparison}\label{apd:aggregation}

We compare three aggregation alternatives at equal edge budget $|C_1|$ (50 Gemma, 20
Llama prompts).
\emph{Stability$\times$influence} (rank union edges by $s(e){\cdot}\mathrm{inf}(e)$)
\emph{hurts} relative to $C_1$ (7/50, 2/20 wins), confirming that re-weighting by
influence re-introduces the greedy bias underlying union-pruned's weakness.
\emph{$\tau$-adaptive} (largest $\tau$ with $\mathrm{IR}(C_\tau)\!\geq\!\mathrm{IR}^*{-}0.02$)
selects $\tau{=}1$ on every prompt: $C_1$ is already within $2\%$ IR of the best individual
config, making strict consensus the data-driven operating point.
A post-hoc oracle (best single config per prompt) does outperform $C_1$ (44/50, 14/20),
but is unavailable without ground-truth per-input knowledge.

\end{document}